\documentclass[11pt]{article}
\usepackage{acl}
\usepackage{times}
\usepackage{latexsym}
\usepackage[utf8]{inputenc}
\usepackage{microtype}
\usepackage{inconsolata}
\usepackage{graphicx}
\usepackage{booktabs}
\usepackage{amsmath}
\usepackage{multirow}
\usepackage{tabularx}
\usepackage{booktabs}
\usepackage{lscape}
\usepackage{adjustbox}
\usepackage{cleveref}
\usepackage{comment}
\usepackage{amsfonts}
\usepackage[tableposition=bottom]{caption}
\usepackage{hyperref}
\usepackage[skins]{tcolorbox}
\title{TRepLiNa: Layer-wise CKA+REPINA Alignment Improves Low-Resource Machine Translation in Aya-23 8B}

\author{
\normalsize 
Toshiki Nakai$^{1}$, Ravi Kiran Chikkala$^{1}$, Lena Sophie Oberkircher$^{1}$, Nicholas Jennings$^{1}$, \\ 
\textbf{\normalsize Natalia Skachkova$^{2}$,   Tatiana Anikina$^{2}$, Jesujoba O. Alabi$^{1}$}\\ 
\textbf{\normalsize } \\
\footnotesize
$^{1}$Saarland University
$^{2}$German Research Center for Artificial Intelligence (DFKI) \\
\texttt{\{toshiki3738,lenaoberkircher\}@gmail.com}\\
\texttt{\{rach00004@teams,s8nijenn@stud,jalabi@cs\}.uni-saarland.de}\\
\vspace{5em}
}

\begin{document}
\maketitle

\begin{abstract}
The 2025 Multimodal Models for Low-Resource Contexts and Social Impact (MMLoSo) Language Challenge targets India’s shortage of resources for low-resource languages (LRLs). The task focuses on building translation systems between High resource languages (HRLs) (Hindi/English) and LRLs (Bhili, Mundari, Santali, and Gondi). Using the MMLoSo 2025 dataset, we investigate whether enforcing cross-lingual similarity in specific internal layers of a decoder-only multilingual large language model (LLM) can improve translation quality from LRLs to HRLs. Specifically, we combine Centered Kernel Alignment (CKA), a similarity metric that encourages representations of different languages to align with Representation Projection Invariance (REPINA), a regularization method that constrains parameter updates to remain close to the pretrained model, into a joint method, we call TRepLiNa (CKA + REPINA). Our results\footnote{we make our code and models public at \url{ https://github.com/konta3738/cka-repina-aya23}} show that aligning mid-level layers with TRepLiNa is a low-cost and practical way to improve LRL translation in data-scarce settings.



\end{abstract}

\section{Introduction}
Many multilingual LLMs share parameters across languages, yet transfer to low-resource languages (LRLs) often lags behind their performance on high-resource languages (HRLs) ~\citep{conneau-etal-2020-unsupervised,zhang-etal-2020-improving}. Recent analysis of Aya-23 8B \citep{aryabumi2024aya}, a multilingual decoder-only model, shows strong neuron overlap across related languages in the embedding layer, perhaps due to token overlap, but it exhibits a marked drop in overlap at intermediate layers \citep{trinley2025languagesdoesaya23think}. Multilingual LLMs tend to turn language-specific inputs into more language-agnostic representations in their middle layers~\citep{wendler-etal-2024-llamas,kojima-etal-2024-multilingual}. This suggests a simple hypothesis: \emph{explicitly encouraging LRL representations to project cleanly into the shared, language-agnostic subspace via cross-lingual similarity metrics should improve transfer to LRLs}. In this project, we focus only on the LRL$\rightarrow$HRL translation. We operationalize this via a lightweight alignment loss between hidden representations of parallel sentences, which is applied at a chosen layer~$\ell$. We use centered kernel alignment (CKA) 
\cite{kornblith2019similarityneuralnetworkrepresentations}, which can robustly compare representations across networks and layers, together with representation projection invariance (REPINA) \citep{razdaibiedina2023representationprojectioninvariancemitigates} to stabilize HRL features against representation drift. We perform experiments, using zero-shot \citep{zhao-etal-2023-pre}, few-shot \citep{karimi-mahabadi-etal-2022-prompt} and QLoRA-based fine-tuning \citep{zhang-etal-2023-machine} on Aya-23~8B, using the MMLoSo benchmark \citep{lrlchallenge2025} pairs, Hindi/English pivots as HRLs; Bhili (Indo-Aryan), Mundari (Austro-asiatic), Santali (Austro-asiatic) and Gondi (Dravidian) as LRLs.

\vspace{0.1 cm}
Our work makes the following \textbf{contributions:}
\begin{itemize}
    \item We present, to the best of our knowledge, the first systematic study of \emph{layer-wise} alignment in a decoder-only LLM for low-resource machine translation (MT), comparing CKA and TRepLiNa (CKA+REPINA) across layers.
    \item We demonstrate that mid-layer alignment (roughly layers 10--15) is most effective, with TRepLiNa consistently favoring layer~15 in limited-data settings.
    \item We show improvements in the weighted composite score of BLEU \citep{papineni-etal-2002-bleu} and ChrF \citep{popovic-2015-chrf}, defined as ($0.6\times$BLEU$+0.4\times$ChrF) with TRepLiNa and provide guidelines on when and where alignment should be applied.
\end{itemize}

\begin{figure*}[t]
    \centering
    \includegraphics[width=0.88\textwidth]{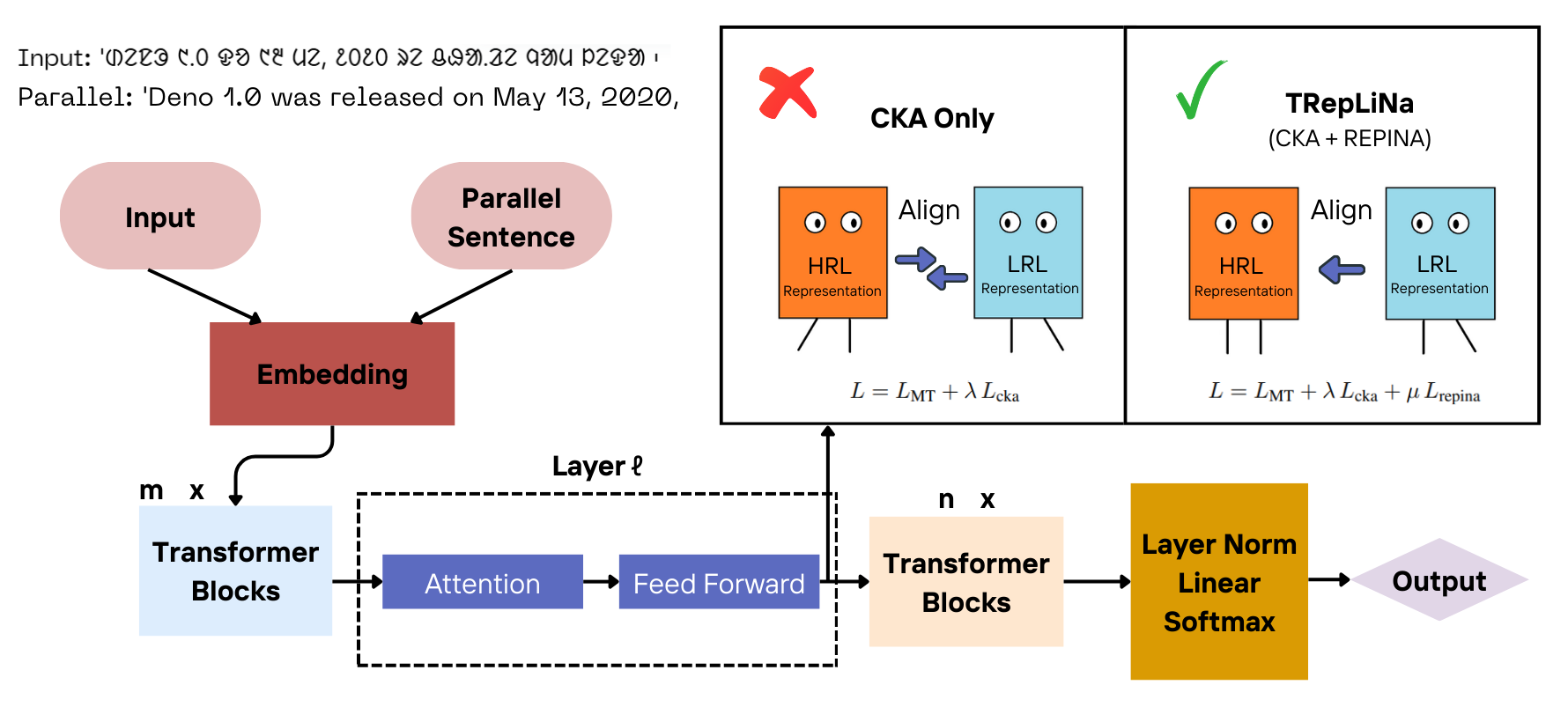}
    \caption{Proposed alignment architecture. Under \textbf{CKA-only}, both HRL and LRL representations drift toward each other, potentially distorting HRL features. By contrast, \textbf{TRepLiNa} constrains HRL representations while guiding LRL representations toward them, achieving targeted alignment without degrading HRL quality. Here, $m$ and $n$ denote the number of transformer blocks before and after the target alignment layer, respectively.}
    \label{fig:RAG}
\end{figure*}

\section{Related Work}
\paragraph{Low-Resource Transfer Methods for Indic LRLs:}
Alongside alignment-based methods, zero-shot and few-shot strategies have also been explored for Indic LRLs. \citet{huidrom-lepage-2020-zero} show that a single multilingual Neural Machine Translation (NMT) model can translate between unseen Indian language pairs, with performance improving as small amounts of parallel data are added. \citet{ghosal-etal-2025-promptrefine} address the problem of improving few-shot generation for Indic LRLs through prompt refinement for MT and other downstream generation tasks. Their findings highlight the importance of designing techniques that enhance low-resource performance. While they focus on input-level prompting, we complement this by aligning hidden representations across layers to improve transfer for Indic LRLs.

\paragraph{Cross-lingual Alignment Methods:}
Cross-lingual alignment has long been studied as a way to enhance transfer in multilingual models, particularly for LRLs~\citep{hammerl-etal-2024-understanding}. Post-hoc cross-lingual alignment methods rotate representations after training, e.g., SVD/orthogonal Procrustes or projection-based removal of language-specific components, improving zero-shot transfer \citep{deb-etal-2023-zero,yang-etal-2021-simple}. Joint optimization injects alignment during training, e.g., cosine-similarity objectives on parallel sentences or contrastive InfoNCE setups \citep{wieting-etal-2019-beyond,pan2021contrastivelearningmanytomanymultilingual} while balancing negatives. CKA has emerged as a computationally attractive alternative to Canonical Correlation Analysis (CCA) \cite{hotelling1936cca} for comparing intermediate activations and for distillation/analysis \citep{dasgupta2025improving}. REPINA \citep{razdaibiedina2023representationprojectioninvariancemitigates} regularizes against representation collapse/drift. We apply these ideas to layer-wise alignment in Aya-23 8B for LRL MT.\footnote{We focus on CKA here; exploring cosine/contrastive or newer similarity objectives (e.g., \citealt{listopad2025wavebasedsemanticmemoryresonancebased}) is left to future work.}

\section{Data}
In this research project, we use the MMLoSo shared task train dataset \citep{lrlchallenge2025} for the experiments with roughly 20k sentence pairs per direction, splitting the dataset into 95\% train and 5\% development. The language pairs include Bhili$\leftrightarrow$Hindi, Mundari$\leftrightarrow$Hindi, Gondi$\leftrightarrow$Hindi (all in Devanagari script) and Santali$\leftrightarrow$English ( Santali in Ol Chiki script and English in Roman script). Our initial tokenization analysis of the data shows that Santali has higher tokenization fertility. It often requires a longer maximum sequence length (368) than Hindi/English (256), which can slightly reduce the tokenwise parallelism available to the alignment loss when sequences must be truncated to apply CKA. 


\section{Methodology and Experiments}
\label{sec:methodology}
In our experiments, we focus on Aya-23 8B, a strong openly available model with broad typological coverage and robust multilingual capabilities. The model is pretrained on 23 languages, including Hindi and English, but it does not cover Mundari, Bhili, Gondi, or Santali. We issue all prompts instructions in English.


\subsection{Prompting}
Here, we discuss the zero-shot and few-shot prompting methods that are used in the experiments.  
\paragraph{Zero-shot:}
In zero-shot experiments, the model relies on its knowledge without any examples \citep{Chikkala2025AutomaticFI}. We consider zero-shot as the baseline for the experiments. See \Cref{fig:zero_shot_prompt} for zero-shot prompt template in the Appendix.

\paragraph{Few-shot:}
In few-shot experiments, we use examples for each language pair from the train set as reference for the language model \citep{anikina2025dfkinit2b}. For each language pair, we use the first example from the training split of the provided data for one-shot, the first three for three-shot, and the first five for five-shot. See \Cref{fig:few_shot_prompt} for few-shot prompt template in the Appendix.

\subsection{TRepLiNa}
This section describes the alignment objective of TRepLiNa. Figure~\ref{fig:RAG} illustrates an overview of our proposed training method.

Given a parallel pair $(x^{(A)}, x^{(B)})$ from an LRL $A$ and a pivot HRL $B$, let $H^{(A)}_{\ell}, H^{(B)}_{\ell}\!\in\!\mathbb{R}^{T\times d}$ denote token wise hidden states at layer $\ell$ (sequence length $T$, width $d$) and $H^{(A)}_{\text{pre} \,\ell}$ be the hidden states obtained from the pretrained model (with an adapter disabled). We augment the MT loss (token-level cross entropy) $L_{\text{MT}}$ with (i) a CKA alignment between LRL/HRL representations and (ii) a REPINA anchoring term that resists drift of HRL features:

{\small
\begin{equation}
L = L_{\text{MT}} + \lambda \, L_{\text{CKA}} + \mu \, L_{\text{REPINA}}
\end{equation}
}
with $\lambda, \mu > 0$. We use linear CKA on mean–centered features:
{\small
\begin{equation}
\begin{aligned}
L_{\text{CKA}} &= 1 - \mathrm{CKA}(H^{(A)}_{\ell}, H^{(B)}_{\ell}), \\
\mathrm{CKA}(H^{(A)}_{\ell}, H^{(B)}_{\ell}) 
&= \frac{\| X^\top Y \|_F^2}
   {\sqrt{\| X^\top X \|_F^2}\,\sqrt{\| Y^\top Y \|_F^2}} .
\end{aligned}
\end{equation}
}
$F$ denotes Frobenius norm. $X$ and $Y$ represent the matrices after applying mean-centering on $H^{(A)}_{\ell}$ and $H^{(B)}_{\ell}$ respectively.
For REPINA, we anchor HRL states to a stop-gradient identity mapping of a reference pass, i.e.,
{\small
\begin{align}
L_{\text{REPINA}}\!\bigl(H^{(A)}_{\text{pre} \,\ell},H^{(A)}_{\ell}\bigr)
\;=\; \bigl\|\,H^{(A)}_{\text{pre} \,\ell} - \tilde{\phi}\!\bigl(H^{(A)}_{\ell}\bigr)\,\bigr\|_2^2,
\end{align}
}
Equivalently, $\tilde{\phi}(\cdot)=\texttt{sg}(\cdot)$; in our implementation this is the detached HRL hidden state at the same layer from the forward pass.
CKA pulls $A$ toward $B$ at layer $\ell$, while REPINA stabilizes $B$. Unless noted, both terms are applied at a single layer $\ell$.

\subsection{Experimental Design}
\paragraph{Step~1: Layer sweep (small data):}
To make the sweep computationally tractable, we sample 1{,}000 parallel pairs and train for one epoch per direction (Mundari $\rightarrow$ Hindi, Santali $\rightarrow$ English). We sweep layers $\ell\!\in\!\{1,2,5,10,15,20,25,30,31,32\}$ and evaluate \textbf{CKA-only} and \textbf{TRepLiNa (CKA+REPINA)} against two baselines \textbf{NoAlign} and \textbf{REPINA-only}.
For \textbf{REPINA-only}, we fix $\ell\!=\!15$ (the best layer observed under TRepLiNa) to isolate the marginal contribution of CKA. We set $\lambda\!=\!\mu\!=\!0.05$, values that are large enough to reveal effects at this data scale, yet small enough to avoid the over-alignment; larger CKA weights (e.g., $\lambda\!=\!0.3$) degraded MT performance in preliminary runs. The \textbf{NoAlign} (standard QLoRA finetuning) excludes both CKA and REPINA.

\paragraph{Step~2: Longer training at the best layer:} Using the best layer from Step~1, we train for up to 5 epochs and track BLEU/ChrF on a 500-sample development set each epoch, comparing TRepLiNa vs. REPINA-only ($\lambda=0.01$, $\mu=0.05$).

\section{Results and Analysis}
Here, we analyze the results of zero-shot, few-shot, TRepLiNa, REPINA and NoAlign from Table \ref{tab:exp2}
\subsection{Step~1: Layer-Wise Trends}
\label{step1results}
\begin{figure}[htbp]
    \centering
    \includegraphics[width=0.45\textwidth]{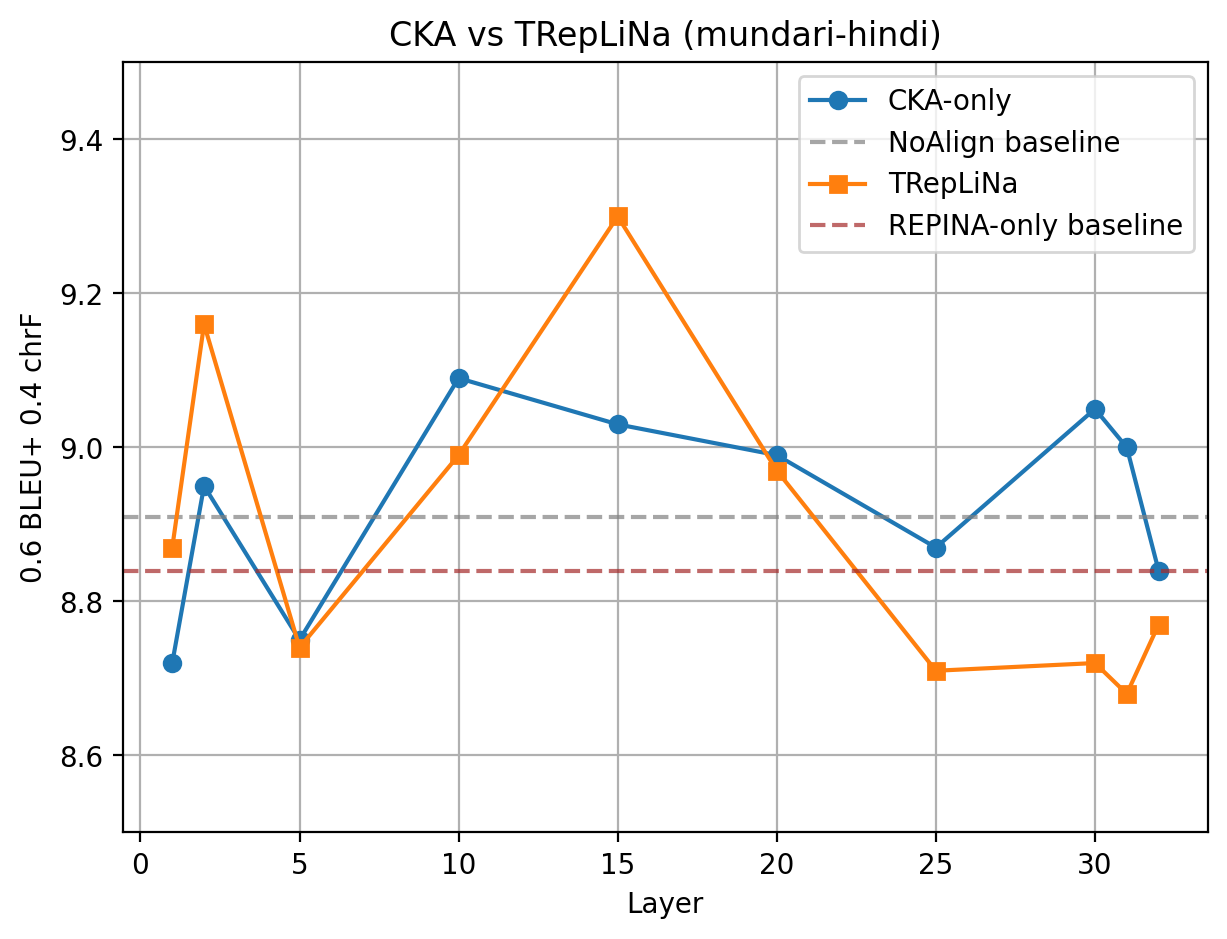}
    \caption{Comparison of ($0.6 \times$ BLEU $+ 0.4 \times$ ChrF) across layers for CKA, REPINA, NoAlign and TRepLiNa.}
    \label{fig:cka_repina}
\end{figure}

The result is discussed for 1k pairs and 1 epoch.
For Mundari--Hindi, the weighted composite score across layers improves (see Figure~\ref{fig:cka_repina}). CKA peaks at layer~10, whereas TRepLiNa peaks at layer~15; the same tendency holds for Santali--English (see Appendix \ref{app:santali-step1}).

\begin{table*}[ht]
\centering
\small
\renewcommand{\arraystretch}{1.0}
\begin{adjustbox}{max width=\textwidth}
\begin{tabular}{lccccccc}
\toprule
\textbf{Language} & \textbf{Zero-shot} & \textbf{Few-shot (1)} & \textbf{Few-shot (3)} & \textbf{Few-shot (5)} & \textbf{TRepLiNa (Ours)} & \textbf{REPINA-only} & \textbf{NoAlign} \\
\midrule
Bhili$\rightarrow$Hindi       &4.75  &4.54  &4.84  &3.96  & 47.96 & \textbf{48.02} & 48.01 \\
Gondi$\rightarrow$Hindi     &4.39  &3.66  &3.75  &3.99  & \textbf{36.26} & 36.18 & 36.25  \\
Mundari$\rightarrow$Hindi     &3.54  &3.00  &3.01  &3.24  & \textbf{34.24} & 33.45 & 33.36 \\
Santali$\rightarrow$English   &1.38  &1.77  &1.05  &1.16  &  \textbf{33.02} & 32.28 & 32.14 \\
\bottomrule
\end{tabular}
\end{adjustbox}
\caption{Final translation scores across language pairs (0.6 × BLEU + 0.4 × ChrF). Best scores are in \textbf{bold.}}
\label{tab:exp2}
\end{table*}

\paragraph{Interpretation:} CKA-only encourages both languages to meet in the middle; without a stabilizer, HRL features may drift, which can blunt gains in later layers. REPINA counteracts this, making mid-high layers (15) the sweet spot when pairing with CKA.

\subsection{Step~2: Multi-Epoch Comparison at Selected Layer}
\label{sec:step2}

\paragraph{Setup:}
Using the best alignment layer from Step~1 (typically a mid-layer around $\ell\!=\!15$), we train for up to five epochs on the full split ($\approx$20k pairs) and evaluate after each epoch on a 500-sample development set. Unless noted otherwise, we set $(\lambda,\mu)=(0.01,0.05)$ for this longer run, i.e., a lower CKA weight than in Step~1 to avoid over-regularization at scale. We report the MMLoSo score (0.6$\times$BLEU + 0.4$\times$ChrF) and also track BLEU/ChrF separately (Appendix Table~\ref{tab:exp2-bleu}). Model selection uses the best development set checkpoint per direction.

\subsection{Findings}

\paragraph{Gondi$\rightarrow$Hindi:} TRepLiNa attains the highest performance score exceeding zero-shot performance.  Few-shot(1) has the lowest score, the gap between the highest and lowest performance scores is 32.6.


\paragraph{Mundari$\rightarrow$Hindi:} TRepLiNa achieves the best score on development set outperforming zero-shot, while few-shot(1) has the lowest score, the difference between the best and the lowest performance score is  29.24.


\paragraph{Santali$\rightarrow$English:} TRepLiNa has the best performance score surpassing zero-shot, whereas few-shot(3) has the lowest score. A difference of 31.97 exists between the best and worst performance scores. For comparison,~\citet{billah2024santalilinguisticinclusionbuilding} report a BLEU of 11.13 on their development set; For comparison, \citet{billah2024santalilinguisticinclusionbuilding} report a BLEU of 11.13 on their development set; our result (Appendix Table~\ref{tab:exp2-bleu}) is 25.24 BLEU, a +14.11 absolute and $\approx$2.27$\times$ relative improvement.


\paragraph{Bhili$\rightarrow$Hindi:} REPINA-only has the highest score, it could be because Bhili and Hindi are typologically close, a strong CKA weight can over-align and wash out beneficial language-specific features. However, our approach TRepLiNa performs better than zero-shot. Few-shot(5) has the lowest score, The highest score exceeds the lowest by 44.06.


\paragraph{Takeaways:}
(i) \emph{Early vs.\ late epochs:} \textbf{NoAlign} shows stronger performance in the initial stages of training with 1k inputs (see Figure \ref{fig:cka_repina}), whereas \textbf{REPINA-only} tends to surpass it when trained on larger datasets (20k).
(ii) \emph{Data scaling:} Larger datasets favor a lower CKA weight; we used $\lambda\!=\!0.05$ for the 1k/1-epoch sweep and $\lambda\!=\!0.01$ for 20k/5-epoch training. As cross-lingual representations become sufficiently aligned, excessive CKA pressure can erode language-specific cues. 
(iii) \emph{Language proximity:} For related pairs (e.g., Bhili--Hindi), We recommend reducing $\lambda$; for more distant pairs, mid-layer TRepLiNa remains robust.

\section{Conclusions}
In this paper, we investigate layer-wise alignment as a simple and effective strategy for improving low-resource translation using Aya-23~8B on MMLoSo language pairs. We show that aligning representations at mid layers enhances performance on translation tasks between language pairs, and that coupling similarity (CKA) with stability (REPINA) in our proposed \textbf{TRepLiNa} method yields robust gains across data-scarce settings. 


%

\section*{Limitations}
We do not explore other similarity objectives (cosine, contrastive InfoNCE) or recent proposals \citep{listopad2025wavebasedsemanticmemoryresonancebased}; we use coefficients ($\lambda$, $\mu$) without scheduler/tuning; and this study does not include a thorough ablation study of the hyperparameters ($\lambda$, $\mu$). In our experiments, we have not explored chain of thought prompting techniques and different prompt templates. From the results Table \ref{tab:exp2}, we observe that there is a reduced performance of TRepLiNa on Bhili$\rightarrow$Hindi, where it underperforms the REPINA-only and NoAlign methods. These results indicate that our method may not generalize well to all language pairs. Santali tokenization sometimes requires longer sequences than 256, reducing token-wise overlap for alignment when truncation occurs. We do not evaluate human adequacy/fluency or domain transfer and qualitative analysis of the generated output by the models.

\section*{Acknowledgments}
We are grateful to our collaborators and mentors for valuable discussions on layer-wise analysis and low-resource MT. We especially thank Mina Abarico for designing the architecture diagram. This project was also supported by the German Federal Ministry of Research, Technology and Space (BMFTR) as part of the project TRAILS (01IW24005), and by Saarland University. We also thank the reviewers for their constructive comments. Any remaining errors are our own.

\bibliography{custom}

\clearpage
\appendix
\section{Appendix: Training and Implementation Details}
\label{sec:appendix}

\subsection{Codebase and Reproducibility}
We provide a single-script trainer for QLoRA fine-tuning of Aya-23 with layer-wise alignment. Seeds are fixed for Python and PyTorch (CPU/GPU). All console/file logs are timestamped; training/eval logs are written via helper functions (\texttt{write\_train\_log}, \texttt{write\_eval\_log}). LoRA adapters are pushed to a Hugging Face repo using access tokens from environment variables.

\subsection{Model, Quantization, and LoRA}
We load \texttt{CohereLabs/aya-23-8B} with 4-bit NF4 (BitsAndBytes) and \texttt{bf16} (or \texttt{fp16}). We enable \texttt{output\_hidden\_states} to obtain intermediate activations. LoRA is applied to standard projection modules \texttt{[q,k,v,o,gate,up,down]} with default $(r{=}16, \alpha{=}32, \text{dropout}{=}0.05)$. We use gradient checkpointing and \texttt{enable\_input\_require\_grads()} to support k-bit training.

\subsection{Tokenization and Batching}
We use a fast tokenizer; if the pad token is missing, EOS is used as PAD. Causal-LM inputs are left-padded; alignment-only passes are right-padded. Prompts follow:
\emph{``Translate to \texttt{\{lang\_b\_name\}}:\textbackslash n\{src\}\textbackslash n''}.
Labels mask the prompt with $-100$. Max lengths are typically 256 (Santali uses 368). We pad to a multiple of 8 for tensor cores. Global batch size is 1 with gradient accumulation (default 16).

\subsection{Data Splits and Development Set}
From a CSV with columns \texttt{src\_col}/\texttt{tgt\_col}, we create train/development set splits. If $=<$1k examples, development set =10\%; otherwise $\approx$5\% (capped 1k–2k). Development set evaluation uses up to 500 examples per epoch.

\subsection{Losses and Layer-wise Alignment (No Equations)}
\paragraph{Task loss:} We use a label-smoothed causal LM loss with $\epsilon=0.1$ over valid target tokens.

\paragraph{Alignment passes (procedure only):} For a parallel pair from LRL~$A$ and HRL~$B$, we:
\begin{enumerate}
    \item Run \emph{source-only} strings for both languages to collect hidden states at a chosen layer $\ell$.
    \item Mask pads, align sequence lengths (truncate to maximum), flatten tokens across the batch, and mean-center features.
    \item Compute a \emph{similarity score} between $A$ and $B$ at layer $\ell$ and add its complement as an alignment penalty.
\end{enumerate}
This is the same CKA objective introduced in the main text; we omit formulas here and refer the reader to the Methodology and Experiments section (Section \ref{sec:methodology}).

\paragraph{REPINA anchoring (procedure only):} Periodically (e.g., every two optimizer steps) we:
\begin{enumerate}
    \item Disable adapters to obtain a \emph{reference} HRL representation at layer $\ell$ on the same inputs.
    \item Penalize the mean-squared deviation between current and reference HRL hidden states (stop-gradient on the reference).
\end{enumerate}
This follows the REPINA scheme described in the main text; equations are intentionally omitted here.

\paragraph{Combined objective:} Training minimizes task loss + similarity penalty + anchoring penalty with user-set coefficients (\texttt{--lambda\_cka}, \texttt{--mu\_repina}). Both terms are applied at a single chosen layer $\ell$.

\begin{figure*}[h!]
    \centering
   \fbox{\includegraphics[width=0.9\textwidth]{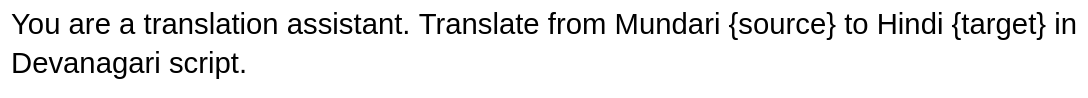}}
    \caption{Zero-shot prompt}
   \label{fig:zero_shot_prompt}
\end{figure*}

 \begin{figure*}[h!]
   \centering
    \fbox{\includegraphics[width=0.9\textwidth]{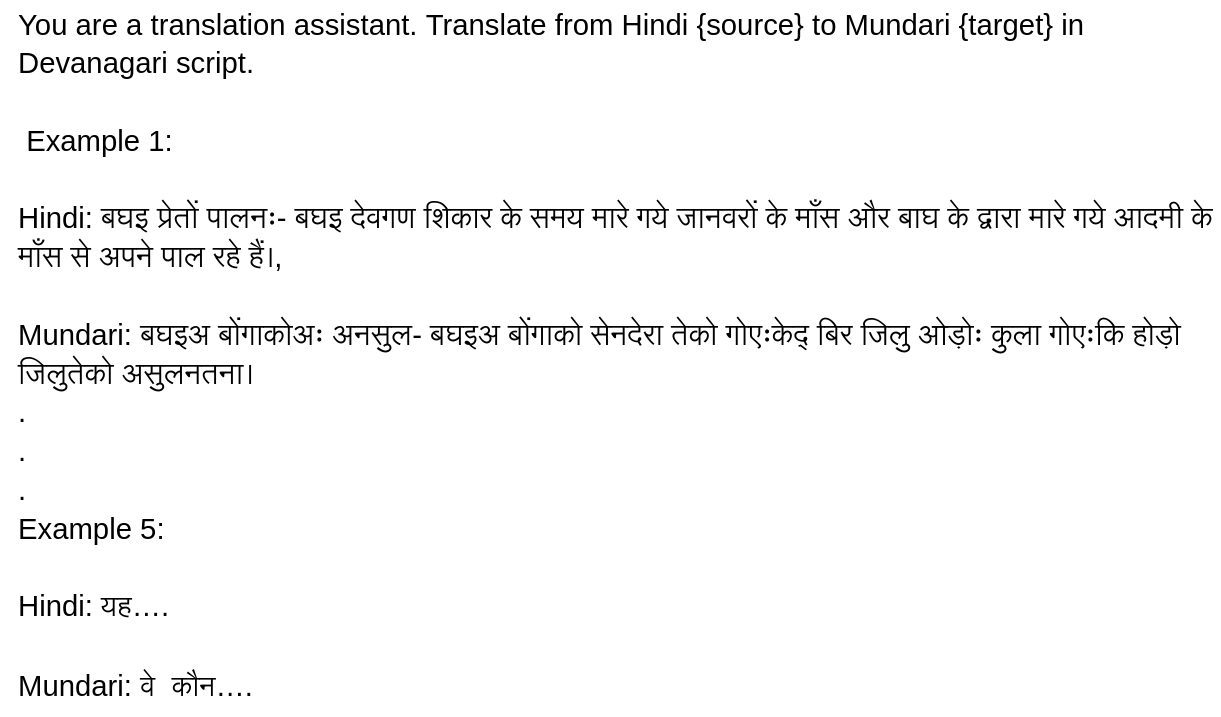}}
   \caption{Few-shot prompt}
   \label{fig:few_shot_prompt}
\end{figure*}

\subsection{Optimization and Precision}
We use PagedAdamW8bit (or AdamW) with $\beta=(0.9,0.95)$, weight decay $0.01$, linear warmup (ratio default $0.05$), and LR in $[1\!\times\!10^{-4}, 2\!\times\!10^{-4}]$ (default $2\!\times\!10^{-4}$). Mixed precision uses \texttt{torch.amp.autocast} (\texttt{bf16}/\texttt{fp16}); for \texttt{fp16}, gradients use \texttt{GradScaler}. We clip global gradients to 1.0 for \texttt{bf16}. Gradients are zeroed with \texttt{set\_to\_none=True}. Optimizer steps occur every \texttt{grad\_accum} micro-steps.

\subsection{Model and Training Defaults}
Unless noted: max source/target 256 (Santali 368), LR $2\!\times\!10^{-4}$, warmup $5\%$, batch size 1, grad accumulation 16, and mixed precision. Layer $\ell$ is selected via sweeps; CKA and REPINA use the same $\ell$.

\subsection{BLEU and ChrF Results (Per Direction)}
\begin{table*}[ht]
\centering
\small
\renewcommand{\arraystretch}{1.2}
\begin{adjustbox}{max width=\textwidth}
\begin{tabular}{lccccccc}
\toprule
\textbf{Language} & \textbf{Zeroshot} & \textbf{Few-shot (1)} & \textbf{Few-shot (3)} & \textbf{Few-shot (5)} & \textbf{TRepLiNa (Ours)} & \textbf{REPINA-only} & \textbf{NoAlign} \\
\midrule
Bhili$\rightarrow$Hindi       & 0.88 & 0.64 & 0.93 & 0.35 & 40.15 & \textbf{40.26} & 40.13 \\
Gondi$\rightarrow$Hindi       &0.37  &0.12  &0.30 &0.56  & \textbf{28.71} & 28.44 & 28.64  \\
Mundari$\rightarrow$Hindi     & 0.14 & 0.06 & 0.04 & 0.08 & \textbf{25.94} & 25.08 & 24.93 \\
Santali$\rightarrow$English   & 0.04 & 0.04 & 0.03 & 0.05 & \textbf{25.24} & 24.64 & 24.26 \\
\bottomrule
\end{tabular}
\end{adjustbox}
\caption{Final translation scores across language pairs (BLEU). Best scores are in \textbf{bold.}}
\label{tab:exp2-bleu}
\end{table*}

\begin{table*}[ht]
\centering
\small
\renewcommand{\arraystretch}{1.2}
\begin{adjustbox}{max width=\textwidth}
\begin{tabular}{lccccccc}
\toprule
\textbf{Language} & \textbf{Zeroshot} & \textbf{Few-shot (1)} & \textbf{Few-shot (3)} & \textbf{Few-shot (5)} & \textbf{TRepLiNa (Ours)} & \textbf{REPINA-only} & \textbf{NoAlign} \\
\midrule
Bhili$\rightarrow$Hindi       & 10.57 & 10.40 & 10.72 & 9.38 & 59.67 & 59.65 & \textbf{59.84} \\
Gondi$\rightarrow$Hindi       &10.42&  8.97 &8.93 &9.12&  47.58 &  \textbf{47.78} & 47.67 \\
Mundari$\rightarrow$Hindi     & 8.66 & 7.43 & 7.48  & 7.98 & \textbf{46.68} & 46.02 & 46.00 \\
Santali$\rightarrow$English   & 3.40 & 4.39 & 2.60  & 2.83 & \textbf{44.68} & 43.74 & 43.96 \\
\bottomrule
\end{tabular}
\end{adjustbox}
\caption{Final translation scores across language pairs (ChrF). Best scores are in \textbf{bold.}}
\label{tab:exp2-chrf}
\end{table*}

\section*{Compute, Runtime, and Practical Notes}
\begin{itemize}
    \item \textbf{Hardware:} Experiments are ran on A100 40GB or H100 80GB (QLoRA fits comfortably); BF16 preferred when available. Training took approximately 30 hours on 1 A100 40GB, and 16 hours on 1 H100 80GB.
    \item \textbf{Stability:} For typologically close pairs (e.g., Bhili–Hindi), reduce the similarity weight over epochs to avoid over-alignment.
    \item \textbf{Layer indexing:} Hidden state tuple index 0 corresponds to the embedding output; a user layer \(\ell\) refers to the 1-based transformer block output.
\end{itemize}

\section{Appendix B: Complementary Results}
\label{sec:appendix-results}


\subsection{Step-1: Layer Sweep on Santali$\rightarrow$English}
\label{app:santali-step1}

\begin{figure}[htbp]
    \centering
    \includegraphics[width=0.45\textwidth]{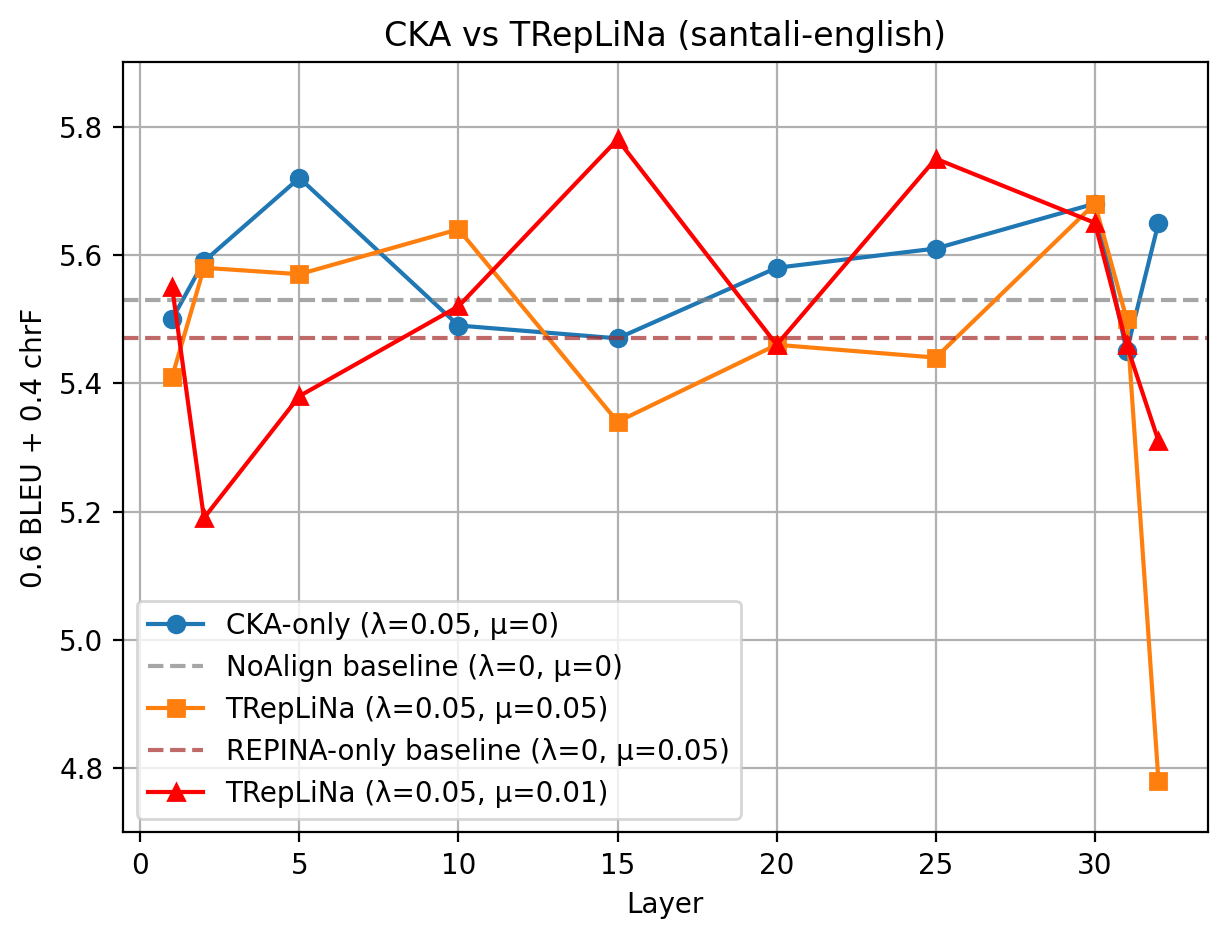}
    \caption{Comparison of $(0.6 \times \mathrm{BLEU} + 0.4 \times \mathrm{ChrF})$ across layers for \textbf{CKA} and \textbf{TRepLiNa} on Santali$\rightarrow$English (1k rows, 1 epoch). Dashed lines indicate each method's baseline.}
    \label{fig:cka_repina_santali}
\end{figure}

With only 1{,}000 training pairs and a single epoch, \textit{anchoring} from REPINA can transiently conflict with task updates: large anchoring (\(\mu\)) tends to pull parameters back toward the reference HRL representation, partially canceling early task learning. Empirically, \(\lambda{=}0.05, \mu{=}0.05\) underperforms \textbf{CKA}-only, but reducing anchoring to \(\mu{=}0.01\) makes \textbf{TRepLiNa} outperform \textbf{CKA}-only. Performance peaks at \(\ell{=}15\), suggesting a mid-layer is most effective for aligning Santali to English in this small-data setting. \textbf{Practical note:} for low data/short training, prefer moderate CKA (\(\lambda{\approx}0.05\)) with lighter anchoring (\(\mu{\approx}0.01\)) and sweep mid-layers (e.g., 10–20).

\subsection{BLEU Table: Summary and Takeaways}
Table~\ref{tab:exp2-bleu} compares final BLEU across settings. On \textbf{Mundari$\rightarrow$Hindi} and \textbf{Santali$\rightarrow$English}, \textbf{TRepLiNa (CKA+REPINA)} achieves the best scores, outperforming both \textit{REPINA-only} and \textit{NoAlign}. For \textbf{Bhili$\rightarrow$Hindi}, \textit{REPINA-only} narrowly leads. few-shot and zero-shot remain far below alignment-based methods, indicating that explicit layer-wise alignment is crucial in the low-resource regime.

\subsection{ChrF Table: Summary and Takeaways}
Table~\ref{tab:exp2-chrf} shows the same comparison in ChrF. The pattern largely mirrors BLEU: \textbf{TRepLiNa} tops \textbf{Mundari$\rightarrow$Hindi} and \textbf{Santali$\rightarrow$English}, while \textit{NoAlign} is slightly best on \textbf{Bhili$\rightarrow$Hindi}. Despite small differences between top systems on Bhili$\rightarrow$Hindi, both metrics agree that alignment generally helps, especially for the more distant pairs. Overall, ChrF confirms the BLEU trends and supports the utility of combining CKA with REPINA.

\section{Appendix C: Future Directions}
\label{sec:appendix-future}

\begin{figure}[htbp]
    \centering
    \includegraphics[width=0.45\textwidth]{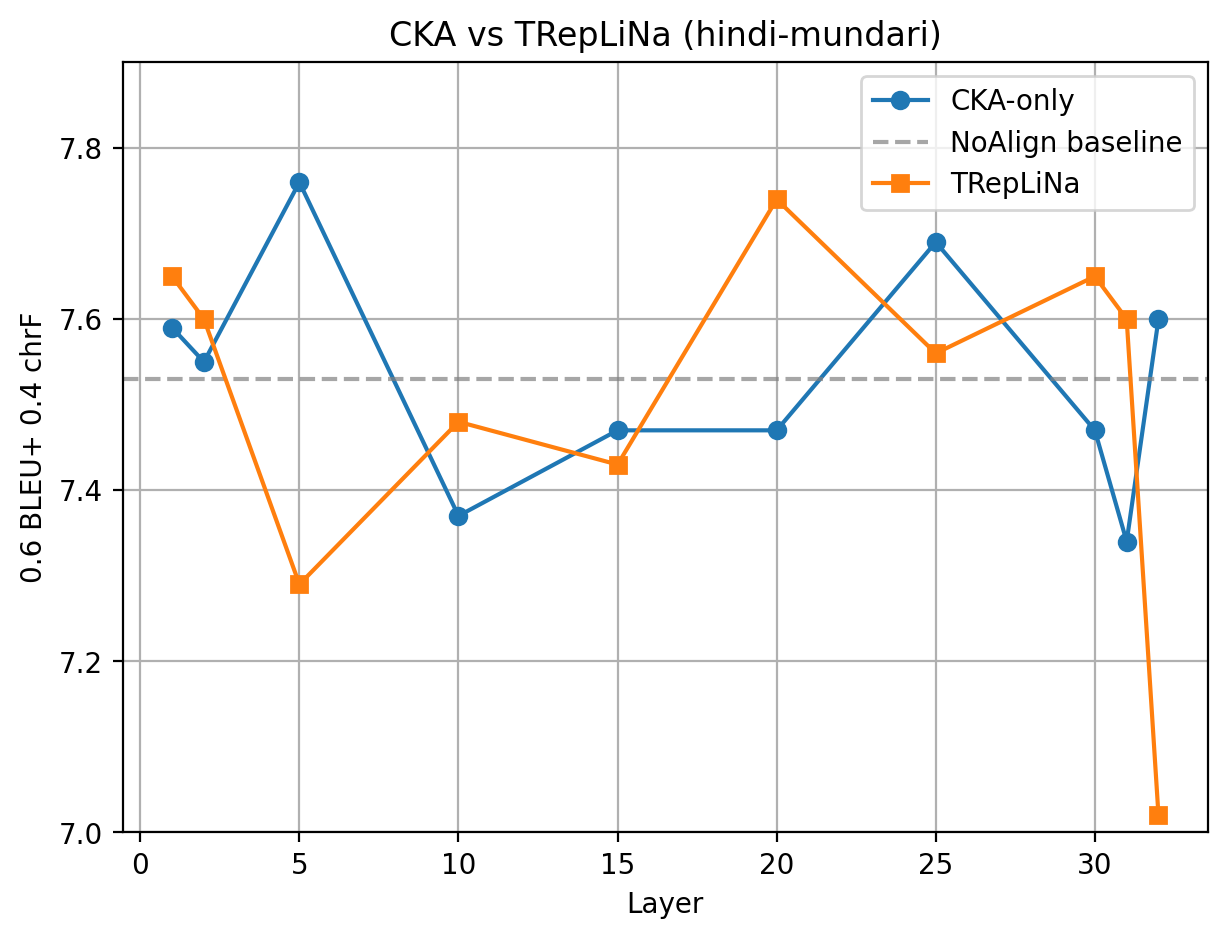}
    \caption{Layer sweep on Hindi$\rightarrow$Mundari (1k pairs, 1 epoch). We plot $0.6\times\mathrm{BLEU}+0.4\times\mathrm{ChrF}$ for \textbf{CKA-only} and \textbf{TRepLiNa}; dashed lines denote each method’s \textsc{NoAlign} baseline. CKA-only peaks at $\ell{=}10$, TRepLiNa at $\ell{=}20$.}
    \label{fig:cka_repina_hindi_mundari}
\end{figure}

\paragraph{Scope:}
We did not explore \textsc{HRL}\,$\rightarrow$\,\textsc{LRL} directions in the main paper due to the asymmetric computational profile of the task and the cost of fine-tuning Aya-23 8B. Here we provide a preliminary \emph{Step-1} layer sweep on Hindi$\rightarrow$Mundari (1k pairs, 1 epoch; $\lambda{=}\,0.05$, $\mu{=}\,0.05$).

\paragraph{Setup and metrics:}
We compare \textbf{CKA-only} and \textbf{TRepLiNa} against the \textbf{NoAlign} baseline across layers, using the combined score $0.6\times\text{BLEU}+0.4\times\text{ChrF}$ (Figure.~\ref{fig:cka_repina_hindi_mundari}).

\paragraph{Observations:}
\textbf{(i)} CKA-only peaks at layer~10 and TRepLiNa peaks at layer~20; both outperform \textsc{NoAlign}.  
\textbf{(ii)} With $\mu{=}\,0.05$ and such a small regime (1k/1 epoch), \textsc{REPINA} can over-regularize, likely dampening short-term task learning. This suggests TRepLiNa may be more competitive under larger budgets (e.g., 20k/5 epochs), where the auxiliary signal has time to synergize with the task objective.

\paragraph{Layer asymmetry:}
For \textsc{LRL}\,$\rightarrow$\,\textsc{HRL}, we observed peaks around layers 10–15 for TRepLiNa, whereas \textsc{HRL}\,$\rightarrow$\,\textsc{LRL} peaks later (layer~20). One plausible explanation is that Aya-23 8B has limited pretrained support for LRL tokens and structures. When the \emph{output} is an LRL (e.g., Mundari), later layers must adapt themselves to generate unseen languages; when the \emph{input} is an LRL, earlier layers need to map LRL signals into language-agnostic features. We leave a rigorous verification of this hypothesis to future work.

Future work may extend this approach to encoder–decoder or speech–text models, and explore adaptive scheduling strategies for alignment strength in truly low-data scenarios.

\end{document}